\documentclass[conference]{IEEEtran}
\IEEEoverridecommandlockouts
\usepackage{cite}
\usepackage{amsmath,amssymb,amsfonts}
\usepackage{algorithmic}
\usepackage{graphicx}
\usepackage{textcomp}
\usepackage{xcolor}
\usepackage{booktabs}
\usepackage{multirow}
\usepackage[normalem]{ulem}
\useunder{\uline}{\ul}{} 

\def\BibTeX{{\rm B\kern-.05em{\sc i\kern-.025em b}\kern-.08em
    T\kern-.1667em\lower.7ex\hbox{E}\kern-.125emX}}
\begin{document}

\title{Low-Bit Integerization of Vision Transformers using Operand Reordering for Efficient Hardware}
\author{
\IEEEauthorblockN{Ching-Yi Lin}
\IEEEauthorblockA{\textit{Department of Electrical and Computer Engineering} \\
\textit{University of Maryland}\\
College Park, USA
}
\and
\IEEEauthorblockN{Sahil Shah}
\IEEEauthorblockA{\textit{Department of Electrical and Computer Engineering} \\
\textit{University of Maryland}\\
College Park, USA
}
% Ching-Yi Lin, and Sahil~Shah
% \thanks{Ching-Yi Lin and S. Shah are with the Department
% of Electrical and Computer Engineering at University of Maryland, College Park, MD, USA e-mail: (sshah389@umd.edu). }% <-this % stops a space}
}
\maketitle
%\title{Conference Paper Title*\\
%\author{Ching-Yi Lin\\
%\IEEEauthorblockA{\textit{Univeristy of Maryland} \\
%\textit{Electrical and Computer Engineering}\\
%College Park, USA \\
%\and
%\IEEEauthorblockN{Sahil Shah}
%\IEEEauthorblockA{\textit{Univeristy of Maryland} \\
%\textit{Electrical and Computer Engineering}\\
%College Park, USA \\
%sshah389@umd.edu or ORCID}
%\and
%\IEEEauthorblockN{3\textsuperscript{rd} Given Name %Surname}
%\IEEEauthorblockA{\textit{dept. name of organization (of %Aff.)} \\
\begin{abstract}

Pre-trained vision transformers have achieved remarkable performance across various visual tasks but suffer from expensive computational and memory costs. While model quantization reduces memory usage by lowering precision, these models still incur significant computational overhead due to the dequantization before matrix operations. In this work, we analyze the computation graph and propose an integerization process based on operation reordering. Specifically, the process delays dequantization until after matrix operations. This enables integerized matrix multiplication and linear module by directly processing the quantized input. To validate our approach, we synthesize the self-attention module of ViT on a systolic array-based hardware. Experimental results show that our low-bit inference reduces per-PE power consumption for linear layer and matrix multiplication, bridging the gap between quantized models and efficient inference.
\end{abstract}


\section{Introduction}
In recent years, Transformers have achieved remarkable success in the natural language processing field, primarily due to their attention mechanism, which enables modeling of dependencies irrespective of their distance in the input or output sequence~\cite{vaswani2017attention}. Inspired by this success, similar concepts have been applied to various computer vision tasks, including image classification {\cite{dosovitskiy2020image}} and object detection{\cite{li2022exploring}}. While Transformer-based models offer strong performance, their success relies on pretraining using large-scale datasets~\cite{dosovitskiy2020image}. One notable limitation is that such pretrained models typically have a fixed architecture—meaning the number of parameters and the computational flow must closely match those of the original model. Consequently, many retraining-based model compression techniques cannot fully leverage the advantages of these pretrained architectures.

To address this challenge, several studies~\cite{li2022q, li2023vit} reduce model size by lowering the precision of weights {in the same model}. Although these approaches improve storage efficiency, they often require dequantizing the low-bit parameter back to floating-point to match the level for subsequent modules during inference~\cite{li2022q}, resulting in no improvement or extra dequantization cost compared to their floating-point counterparts.

In this work, we propose a low-bit precision approach that builds upon previous quantization successes while explicitly optimizing computation during model inference. Specifically, our framework ensures both low-bit model storage and a specialized architecture that directly employs low-bit weights and activations. Furthermore, to benefit from the superior accuracy of existing pretrained models, our method is built upon publicly available checkpoints, extending the integerization recipe to other applications without sacrificing the advantages of pretrained Transformer-based models.

\section{Background}

% \subsection{Visual Transformer}

% VisualOne key contribution of ViT is to demonstrate that finetuning a model pretrained from a larger dataset results in a more accurate model than training from scratch. In this work, we follow this concept

\subsection{Model Quantization}

Model quantization has been widely studied as a method to reduce the storage requirements by lowering the precision of weight and activation. Various approaches have been proposed to represent floating-point weights and activations using low-bit precision, including power-of-two quantization~\cite{lin2021fq}, differentiable quantization~\cite{li2022q}, and twin uniform quantization~\cite{yuan2022ptq4vit} in Visual Transformer (ViT) model.

Despite the reduction in memory offered by these techniques, they often require de-quantization before the computationally-heavy operations, resulting in no significant improvement in inference performance. Some studies have proposed quantization schemes that enforce the integer-only outputs~\cite{jacob2018quantization}; however, this approach limits expressivity and typically restricts quantization to only 8-bit precision.

In this work, rather than applying integerization during both training and inference phase, we build on the success of low-bit quantized models and introduce a post-integerization method. This method enables the lower-bit precision integerization models while preserving their expressivity and accuracy.

\subsection{Visual Transformer Hardware Implementation}

With the success of ViT in computer vision tasks, both research groups and companies have begun developing hardware implementations to enhance its performance. Although there are several implementation focused on general-purpose processing engine (PE)~\cite{moon2025t}, we focus on customized datapath. A customized datapath enables minimizing memory access and it also successfully reduces power and latency in commercial hardware~\cite{jouppi2017datacenter,li2020ftrans}. For ViTs, Nag~\cite{nag2023vita} built systolic arrays for efficient matrix operations, and Huang~\cite{huang2023integer} further proposed a group-systolic array to balance power and area efficiency compared to the classic systolic array.

\section{Model Integerization}
In this work, we propose an approach to integerize the model by reordering operations so that certain blocks receive integer inputs. Since the floating-point weights and inputs are de-quantized from low-bit weights and activation, our primary objective is to delay dequantization until after computationally intensive steps—such as linear layers or matrix multiplications. By doing so, we can implement these operations as efficient low-bit multiply-accumulate (MAC) operations, followed by precise post-dequantization. In some cases, this post-scaling can even be absorbed into subsequent operations, resulting in a series of low-bit operations that remains equivalent to the original floating-point formulation. Although we focus on integerizing the self-attention module, the same principles can be extended to other components of ViT.

\begin{figure}
    \centering
    \includegraphics[width=0.9\linewidth]{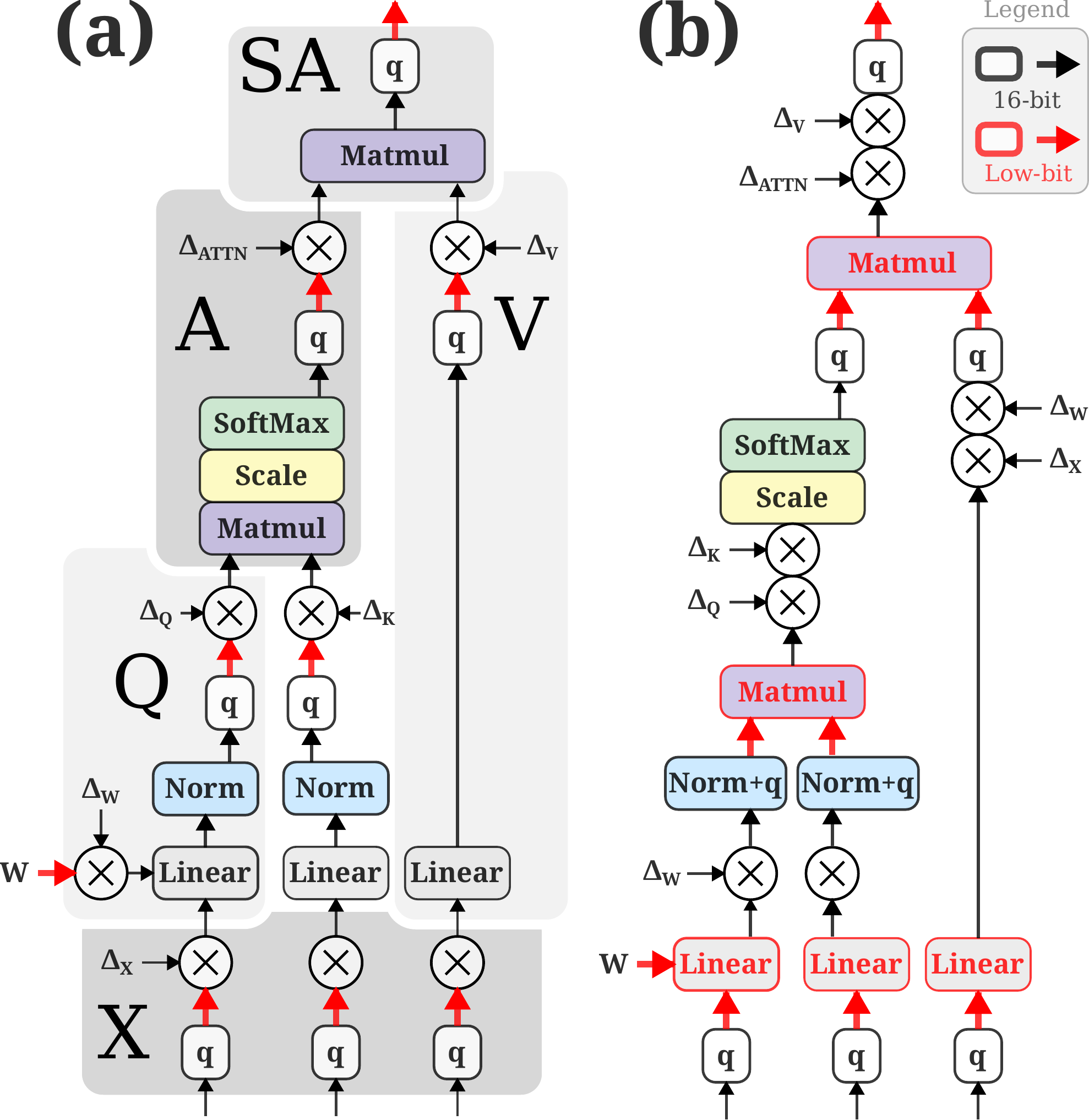}
    \caption{Comparison of inference paths between (a) Q-ViT~\cite{li2022q} and (b) our proposed model (with low-bit precision operation highlighted in red). In Q-ViT, weights and activations must be de-quantized before being processed by linear layers and matrix multiplications. Our proposed approach reorders operations to enable direct low-bit computation in linear layers and matrix multiplications, thereby improving inference performance.}
    \label{fig:integerization}
    \vspace{-4mm}
\end{figure}

\subsection{Integerization Through Operation Reordering}
To integerize the self-attention module of ViTs, we first construct a datapath graph of its operations. Figure \ref{fig:integerization}(a) illustrates the inference path of Q-ViT~\cite{li2022q}, a quantized but not integerized model, which consists of linear layer, layer normalization, matrix multiplication, scale, and softmax. Although the quantizer (q) reduces inputs to fewer bits, all low-bit values are subsequently de-quantized by floating-point multipliers before reaching the computational units. To emphasize this, we mark the low-bit datapath in red, highlighting the absence of direct low-bit inputs in key modules.

In contrast, Fig. \ref{fig:integerization}(b) shows the datapath of our integerized ViT. Here, the scaling units are repositioned to follow computational unit, enabling linear layer and matrix multiplication to be executed in low-bit precision. While layer normalization, scaling, and softmax remain in {full precision}, they are computationally less demanding ($O(N^2)$) compared with the linear layer and matrix multiplication ($O(N^3)$).

\section{Hardware Implementation}

\begin{figure}
    \centering
    \includegraphics[width=0.9\linewidth]{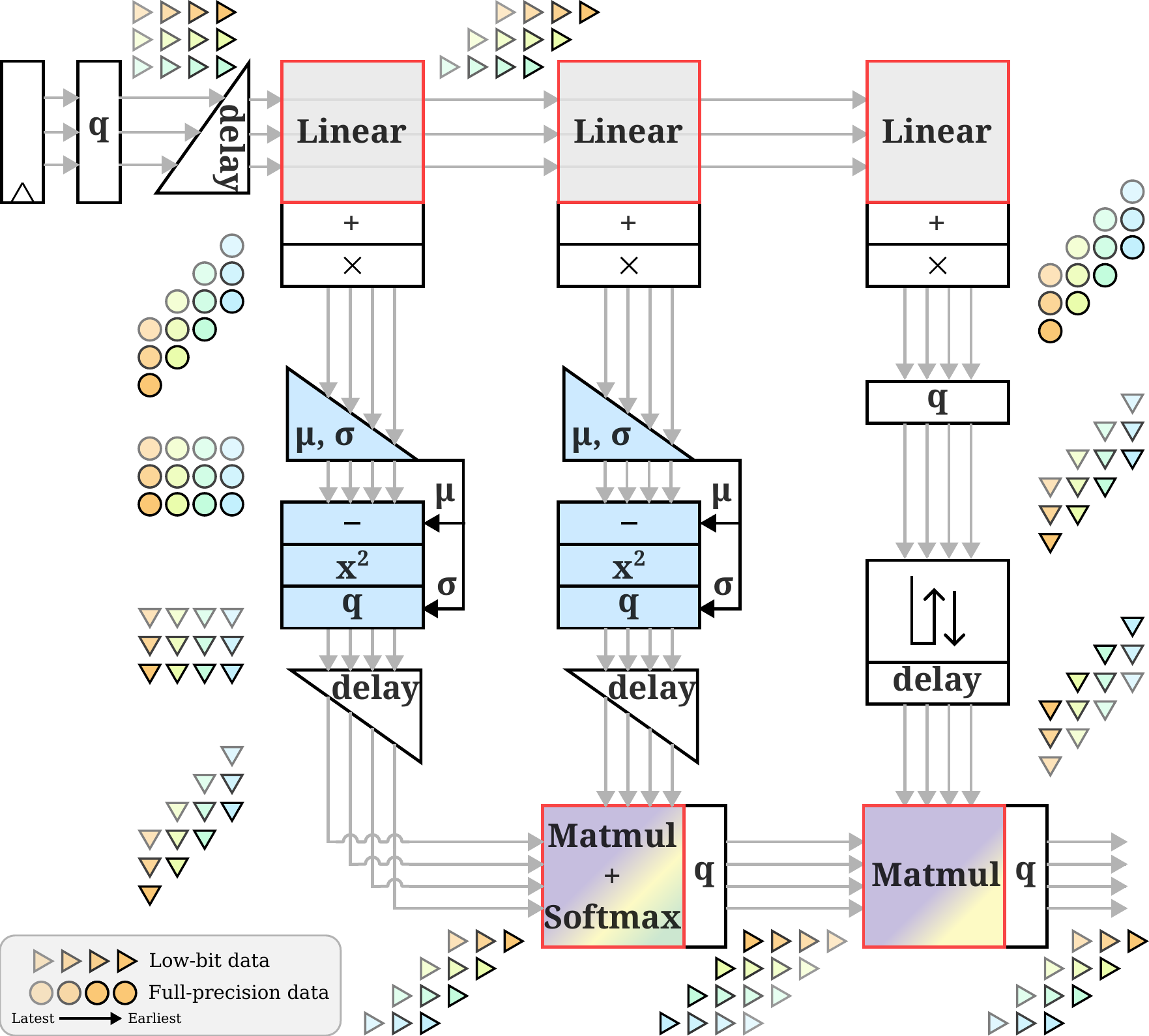}
    \caption{Overview of key modules and dataflows in our proposed hardware architecture. Quantized data are represented by triangle symbols, while full-precision data are depicted as circles. The transparency of these symbols encodes time information.}
    \label{fig:hardware}
    \vspace{-4mm}
\end{figure}

To validate our approach and demonstrate its effectiveness, we design hardware that implements the integerized self-attention module. The key components and dataflows are shown in Figure \ref{fig:hardware}. The main modules include linear layer, layer normalization (LayerNorm), matrix multiplication, matrix multiplication with embedded softmax, a reversing module, and multiple quantizers {(q)}. Dataflow between modules is depicted using triangles (low-bit) or circles (full precision), with channel and time information encoded in color and transparency.

In the following sections, we will describe the detailed implementation of each module.
\subsection{Low-bit Systolic Array for Linear Layer}

The quantized linear layer can be expressed as the product of dequantized input ({${\mathbf{X_{q}}}$}) and weight ({${\mathbf{W_{q}}}$) with the dequantization achieved by multipliers as shown in Figure \ref{fig:integerization} (a) with individual step size vector $\Delta_{\mathbf{X}}$ and $\Delta_{\mathbf{W}}$. This linear layer can be written formally as
\begin{equation}
\label{eqn:linear-fp}
\begin{split}
    \mathbf{Y} &= [\mathbf{X_{q}}(diag(\Delta_{\mathbf{X}}))][\mathbf{W_{q}^T}(diag(\Delta_{\mathbf{W}}))] + \mathbf{b} \\
    &=[ \mathbf{X_{q}}diag(\Delta_{\mathbf{X}})\mathbf{W_{q}}^T]diag(\Delta_{\mathbf{W}}) + \mathbf{b}
\end{split}
\end{equation}
where the channel-wise dequantization achieved by multiplying with diagnoal matrix built from vector-to-matrix diag operator
$diag(\vec{a})$ = $diag(a_1, a_2, \dots, a_n)$
% = $(d_{i,j})_{1\leq i,j\leq n}$$\begin{cases}
%     a_i, \text{if }i=j \\
%     0, \text{else}
% \end{cases}$
= $\begin{bmatrix}
a_1 & 0 & \dots & 0\\
\vdots & \ddots & \ddots & \vdots \\
0  & \dots & 0 & a_n
\end{bmatrix}$.

To facilitate computation in linear layer using lower-bit precision, we replace the channel-wise input scale $\Delta_\mathbf{X}$ with a single $\overline{\Delta_\mathbf{X}}$. This further simplify the processing as follows
\begin{equation}
\label{eqn:linear-3b}
\begin{split}
    \mathbf{Y} &\approx [\mathbf{X_{q}}(\overline{\Delta_{\mathbf{X}}}\mathbf{I})\mathbf{W_{q}}^T]diag(\Delta_{\mathbf{W}}) + \mathbf{b}diag(\frac{\Delta_{\mathbf{W}}}{\Delta_{\mathbf{W}}}) \\
    &= [\mathbf{X_{q}}\mathbf{W_{q}}^T +\frac{\mathbf{b}}{\overline{\Delta_{\mathbf{X}}}}diag(\frac{1}{\Delta_{\mathbf{W}}}) ](\overline{\Delta_{\mathbf{X}}})diag(\Delta_{\mathbf{W}})
\end{split}
\end{equation}

{Equation \ref{eqn:linear-3b}} can be achieved, as shown in Figure \ref{fig:integerization} (b),  with a low-bit linear layer $(\mathbf{X_{q}}\mathbf{W_{q}}^T)$, an equivalent bias term $(\mathbf{b}(\overline{\Delta_{\mathbf{X}}})^{-1}diag(\frac{1}{\Delta_{\mathbf{W}}}))$ and a post-scaling factor $(diag(\Delta_{\mathbf{W}}))$ with $(\overline{\Delta_{\mathbf{X}}})$ canceled by the subsequent LayerNorm.
% The scalar $\overline{\Delta_{\mathbf{X}}}$ can be further cancelled out through the subsequent LayerNorm, resulting a channel-wise multiplier $\Delta_{\mathbf{W}}$ in Figure \ref{fig:integerization} (b).

\subsection{Matrix Multiplication with on-PE Softmax}

\begin{figure}
    \centering
    \includegraphics[width=0.9\linewidth]{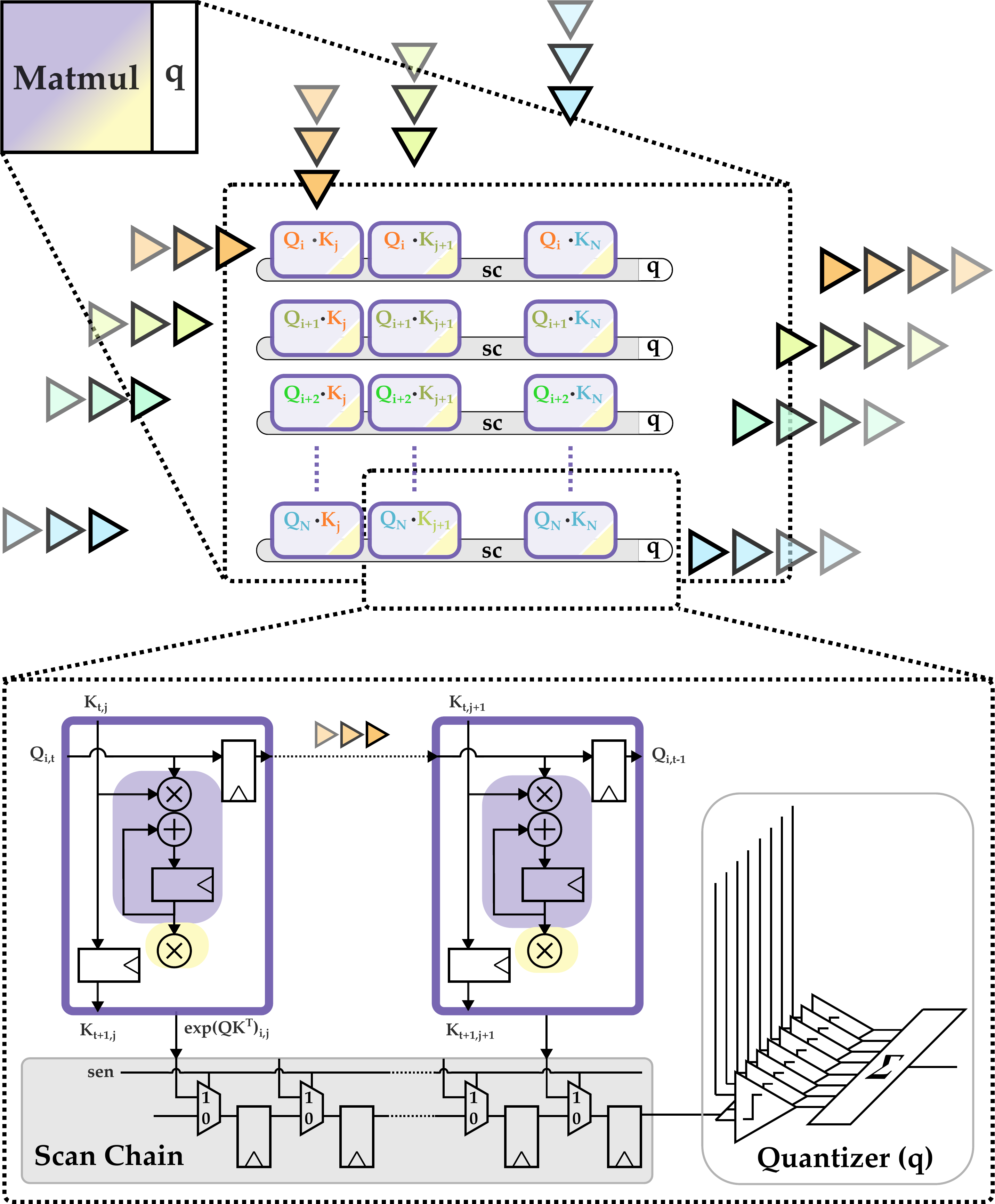}
    \caption{Implementation of matrix multiplication using a systolic array with a scan chain dedicated for each row. Each Processing Element (PE) at position $i,j$ calculates the partial sum of ${\mathbf{QK}^T}_{i,j}$ with a local low-bit MAC unit. The accumulated result is pushed into the scan chain once all operands are processed. The scan chain sequentially outputs the results to the quantizer, generating low-bit output for next module}
    \label{fig:matmul_attn}
    \vspace{-4mm}
\end{figure}

Matrix multiplication in ViTs is used for both the pre-softmax transformation ($\mathbf{QK^T}$) and the weighted sum of value ($\mathbf{W}_{attn}\mathbf{V}$). Although both operations rely on matrix multiplication {and can be realized in a weight stationary manner~\cite{chen2016eyeriss}}, they differ in implementation based on subsequent processing.

For $\mathbf{W}_{attn}\mathbf{V}$, since this matrix multiplication result is passed onto a quantizer, it can be performed at lower bit precision by absorbing the input scales for both operands within the quantizer{, similar to Equation \ref{eqn:linear-3b}}. We implement this matrix multiplication using a systolic array, as shown in Figure \ref{fig:matmul_attn}: Two input matrices are streamed channel-wise, and each PE contains a local low-bit MAC unit. Final results are latched into a scan chain and pushed out sequentially, while the quantizer is realized with a parallel comparator and an adder.
% One noting thing is because we force the data flowing as the same direction as the input to prevent the congestion, the scan chain output has the reversed order as the input. This is visualized by the transparency order in the Fig. xxx

In contrast, the pre-softmax transformation ($\mathbf{QK^T}$) requires scaling and feeding into the softmax module:
\begin{equation}
\label{eqn:attn}
\begin{split}
    attn_{i,j} &= softmax(sQ_{(i,:)}K_{(:,j)})\\
     &= exp(sQ_{(i,:)}K_{(:,j)}) / \Sigma_{j} exp(sQ_{(i,:)}K_{(:,j)})\\
\end{split}
\end{equation}

To simplify the exponential operation, we approximate $e^x$ using base-2 exponentiation and integer shift~{\cite{li2023vit}}
\begin{equation}
\label{eqn:exp}
\begin{split}
    &exp(sQ_{(i,:)}K_{(:,j)}) \\
    % &=(e^s)\wedge (Q_{(i,:)}K_{(:,j)})\\
    &=2\wedge (s \times log2(e) (Q_{(i,:)}K_{(:,j)})) \\
    &=2\wedge (\lfloor s \times log2(e) (Q_{(i,:)}K_{(:,j)})\rfloor + r)\\
    % &=(1 << (\lfloor s \times log2(e) (Q_{(i,:)}K_{(:,j)})\rfloor)\times 2^r \\
    &=2^r << (\lfloor s \times log2(e) (Q_{(i,:)}K_{(:,j)})\rfloor \\
    &\approx(r+1) << (\lfloor s \times log2(e) (Q_{(i,:)}K_{(:,j)})\rfloor \\
\end{split}
\end{equation}

where $r = s \times log2(e) (Q_{(i,:)}K_{(:,j)}) - \lfloor s \times log2(e) (Q_{(i,:)}K_{(:,j)})\rfloor$ is the residual part of the exponent-2.

% For the division part in softmax, we apply the same absorption trick as in the weighted sum quantizer. Rather then dividing by $\Sigma_{j} exp(sQ_{(i,:)}K_{(:,j)})$, we instead scale the quantization step size $\Delta_{ATTN}$ by this sum. 

% Here, the exponent is split into integer part $\lfloor s \times log2(e) (Q_{(i,:)}K_{(:,j)})\rfloor$ and residual part $r\in[0,1)$. The residual part $r$ can be estimated by $r+1$, and the integer part is equivalent to its shift.

% With the exponential logic, the softmax quantizer can be implemented with the same tricks as weighted sum quantizer, that is, absorb the same term in quantizer. In other words, instead of comparing a ratio $exp(sQ_{(i,:)}K_{(:,j)}) / \Sigma_{j} exp(sQ_{(i,:)}K_{(:,j)})$ to step size $\Delta_{ATTN}$, we compare the numerator term $exp(sQ_{(i,:)}K_{(:,j)})$ to the step size scaled by the row exponential sum $\Delta_{ATTN}\times\Sigma_{j} exp(sQ_{(i,:)}K_{(:,j)})$. Although this step size is determined after all the exponential is calculated, its value is fixed across the same row.
Figure \ref{fig:matmul} demonstrates the hardware to realize this function. The output of matrix multiplication (purple) is passed through the scaled exponential unit. While storing results in the scan chain, the factor $\Sigma_{j} exp(sQ_{(i,:)}K_{(:,j)})$ is calculated and propagated to the end of the row. This sum is fed into the quantizer, built from a series of multiplier with boundary values ($-3.5\Delta_{ATTN}$, ..., $1.5\Delta_{ATTN}$, $2.5\Delta_{ATTN}$ in 3-b example) as the comparator reference.

\begin{figure}
    \centering
    \includegraphics[width=0.9\linewidth]{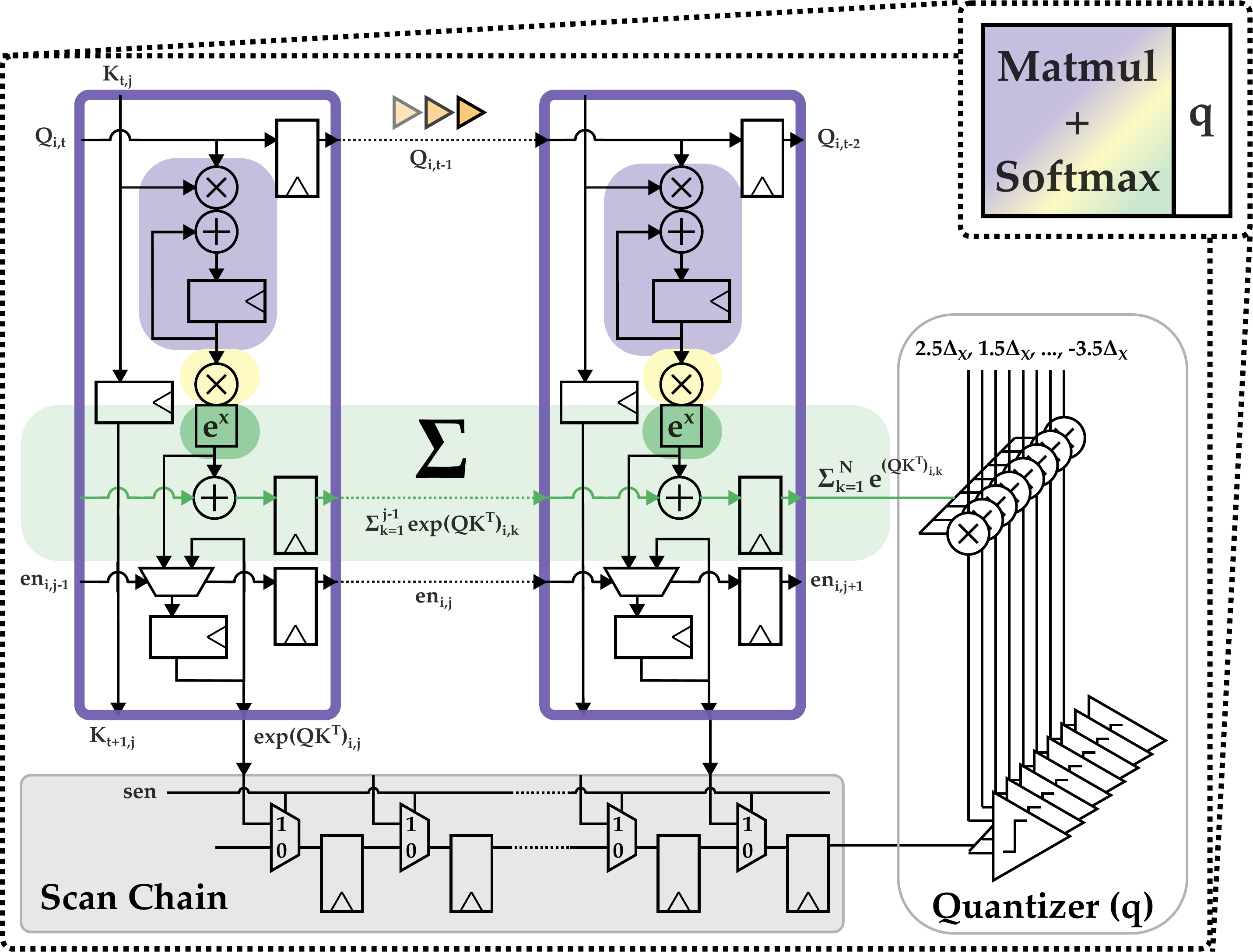}
    \caption{
    Implementation of matrix multiplication with embedded softmax: The design integrates exponential logic and systolic adders within each PE to compute the exponentials of MAC results and accumulate ($\Sigma$) their partial sums. The scan chain propagates exponential results similar as Figure \ref{fig:matmul} with a quantizer scaled by the exponential sum.}
    \label{fig:matmul}
    \vspace{-6mm}
\end{figure}

\subsection{Systolic-compatible LayerNorm}

Another module not in traditional neural network is LayerNorm. This module normalizes each row to $\mathcal{N}(\beta, \gamma^2)$ before feeding it into a quantizer with reference $s_Q = (k - \frac{1}{2})\Delta_Q$, $k=-2^{nbit-1}, ..., 2^{nbit-1} - 1, 2^{nbit-1}$. However, computing mean and variance is challenging for hardware due to the division and square root operation.

The division can be easily converted into reference scaling as the previous absorption trick. We also avoid square root by using variance $\sigma^2$ directly in the comparison logic with additional sign logic for its correctness. These tricks result in the square root and division-free comparator in Figure \ref{fig:comparator}.

For mean and variance computation, we follow the incremental statistics
% Similar to other quantizer, the division at input can be converted to multiplication at comparator reference. On the other side, to get rid of square root in standard deviation $\sigma$, we instead use variation $\sigma^2$ during comparison, with extra criteria about the sign. The mean and variation in systolic array can be calculated from Equation \ref{eqn:running-stat}. Combine both tricks together, the 1-bit quantizer can be realized using Equation \ref{eqn:normq}.
\begin{equation}
\label{eqn:running-stat}
\begin{split}
    \begin{cases}
        \mu_i = \mu_{i-1} + \frac{x_i - \mu_{i-1}}{i}\\
        \sigma^2_i = \sigma^2_{i-1} + (x_i-\mu_{i-1})(x_i-\mu_{i})
    \end{cases}
\end{split}
\end{equation}

with initial condition $\mu_0 = 0$ and $\sigma^2_0 = 0$. Equation \ref{eqn:running-stat} can be realized in a systolic dataflow with a $\mu$ row and a $\sigma^2$ row of PE {attached at the array}{, with results {sent} to the comparator array in Figure \ref{fig:comparator} for the complete} pre-LayerNorm quantization.

% \begin{equation}
% \label{eqn:normq}
% \begin{split}
%     &\frac{x-\mu}{\sigma}\times\gamma+\beta > s\\
%     &\equiv x-\mu > (s-\beta) \times \sigma/\gamma\\
%     &\equiv
%     \begin{cases}
%         \|x-\mu\| > \|(s-\beta) \times \sigma/\gamma\|, \\
%         \hspace{21mm}\text{for }x-\mu >0 \text{ and }(s-\beta) \times \sigma/\gamma > 0\\
%         True,  \hspace{11mm}\text{for }x-\mu >0 \text{ and }(s-\beta) \times \sigma/\gamma > 0 \\
%         False,  \hspace{10mm}\text{for }x-\mu >0 \text{ and }(s-\beta) \times \sigma/\gamma > 0 \\
%         \|x-\mu\| < \|(s-\beta) \times \sigma/\gamma\|, \\  \hspace{21mm}\text{for }x-\mu >0 \text{ and} (s-\beta) \times \sigma/\gamma > 0
%     \end{cases}
% \end{split}
% \end{equation}

\begin{figure}
    \centering
    \includegraphics[width=0.9\linewidth]{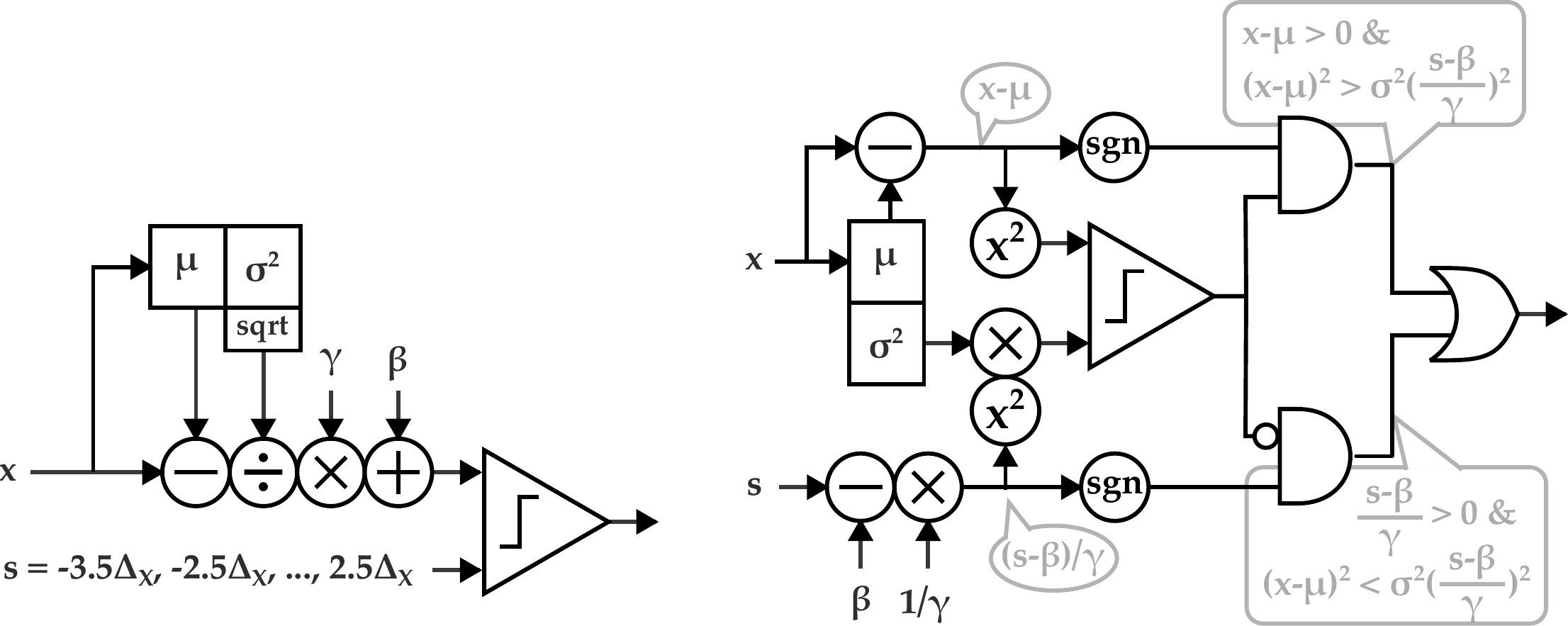}
    \caption{Post-quantized layerNorm implementation: (a) The direct implementation ($\frac{x-\mu}{\sigma}\times\gamma+b > s$) (b) Division and square root-free implementation, replacing $\sigma$ division with $\sigma^2$ multiplication and incorporating sign logic (sgn). The comparator computes $(x-\mu)^2 > \sigma^2 \times[(x-\beta)\times\frac{1}{\gamma}]^2$ instead of the direct formula}
    \label{fig:comparator}
    \vspace{-4mm}
\end{figure}

\section{Experimental Results}
\subsection{Datasets and Training Details}

\textbf{Datasets.} Our experiments are conducted on the CIFAR-10 dataset, which consists of 10 classes and 50k training images, and 10k validation images. We apply the same data augmentation techniques as described in DeiT~\cite{touvron2021training}.

\textbf{Models. }Using a model pretrained from a larger dataset has been shown to achieve higher accuracy compared to training from scratch~\cite{dosovitskiy2020image}. Therefore, in this work, we utilize DeiT-S \cite{touvron2021training} pretrained by {Meta}. This model is distilled from a larger pretrained model on ImageNet-1k, a dataset comprising 1.28 million training images and 1,000 object classes.

\textbf{Experimental settings.} In our experiments, we initialize the weights using DeiT-S model \cite{touvron2021training} and train the model in two phases: the last-layer phase and the fine-tuning phase. Both phases use a base learning rate 5e-4, a batch size 32, the LAMB~\cite{you2019large} optimizer without weight decay, and a cosine annealing scheduler for 300 epochs. The key difference is that the last-layer phase only trains the last layer while the fine-tuning phase trains all the layers.

\subsection{Power Analysis}
One of the key benefits for low-bit inference is the lower number of bits in MAC operations. To validate this claim, we synthesize our implementation on AMD Spartan™ 7 FPGAs at a clock rate 100 MHz in 3-bit resolution. Table \ref{tab:power} presents the power of the main blocks within a self-attention module.

An observation from Table \ref{tab:power} is that the linear layers and matrix multiplication dominate both the number of operations (OPs) and power consumption. Additionally, despite their high computational load, these two blocks exhibit lower power consumption per PE compared to other blocks. These findings demonstrate that our implementation effectively reduces the power in the most computationally intensive units.

\begin{table}[]
\begin{tabular}{@{}ccccccc@{}}
\toprule
 &  & \multicolumn{2}{c}{\# of PE} & \multirow{2}{*}{\begin{tabular}[c]{@{}c@{}}\# of MAC\\ (M)\end{tabular}} & \multicolumn{2}{c}{Power} \\
 &  &  & Actual &  & \begin{tabular}[c]{@{}c@{}}Total\\ (W)\end{tabular} & \begin{tabular}[c]{@{}c@{}}Per PE\\ (mW)\end{tabular} \\ \midrule
\multirow{3}{*}{Q} & Linear & I$\times$O & 24,576 & 4.87 & 10.188 & 0.414 \\
 & LayerNorm & 2$\times$O & 128 & 0.03 & 0.598 & 4.67 \\
 & delay & N$\times$O & 12,672 & {-} & 0.858 & {-} \\ \midrule
\multirow{3}{*}{K} & Linear & I$\times$O & 24,576 & 4.87 & 10.188 & 0.414 \\
 & LayerNorm & 2$\times$O & 128 & 0.03 & 0.598 & 4.67 \\
 & delay & N$\times$O & 12,672 & {-} & 0.858 & {-} \\ \midrule
\multirow{2}{*}{V} & Linear & I$\times$O & 24,576 & 4.87 & 10.399 & 0.423 \\
 & reversing & O$\times$O & {-} & {-} & 1.511 & {-} \\ \midrule
 $\mathbf{QK^T}$ & \begin{tabular}[c]{@{}c@{}}Matmul\\ + softmax\end{tabular} & N$\times$N & 39,204 & 2.51 & 58.959 & 1.504 \\ \midrule
 $\mathbf{PV} $ & Matmul & N$\times$O & 12,672 & 2.51 & 4.597 & 0.362 \\ \midrule
 &  &  &  &  &  & 
\end{tabular}
\caption{Power consumption of primary blocks in 3-bit self-attention}
\label{tab:power}
    \vspace{-7mm}
\end{table}

\subsection{Model Comparison}

To assess the performance of our integerized model, we compare it against other approaches using DeiT-S \cite{touvron2021training} architecture, as shown in Table \label{tab:acc}. Both I-BERT~\cite{kim2021bert} and I-ViT~\cite{li2023vit} support integer-only operations, but their weights and activations are limited to 8 bits. In contrast, Q-ViT~\cite{li2022q} quantizes the model to 2 bits and 3 bits; however, its model inference requires weights and activations to be de-quantized to floating-point format before being processed in linear layer and matrix multiplication.

\begin{table}[b]
\begin{tabular}{@{}ccccccc@{}}
\toprule
 & \multirow{2}{*}{Int-only} & \multicolumn{2}{c}{Parameter} & \multirow{2}{*}{\begin{tabular}[c]{@{}c@{}}OPs\\ (G)\end{tabular}} & \multirow{2}{*}{Multiplier} & \multirow{2}{*}{\begin{tabular}[c]{@{}c@{}}CIFAR-10\\ Accuracy\end{tabular}} \\
 &  & \begin{tabular}[c]{@{}c@{}}Number\\ (M)\end{tabular} & \begin{tabular}[c]{@{}c@{}}Size\\ (MB)\end{tabular} &  &  &  \\ \midrule
I-BERT~\cite{kim2021bert} & V & \multirow{6}{*}{21.8} & 21.8 & \multirow{6}{*}{4.3} & INT8 & - \\
I-ViT~\cite{li2023vit} & V &  & 21.8 &  & INT8 & - \\
\multirow{2}{*}{Q-ViT~\cite{li2022q}} & \multirow{2}{*}{X} &  &  5.8 &  & FP32 & \textbf{93.91} \\
 &  &  & 8.3 &  & FP32 & \textbf{97.04} \\
\multirow{2}{*}{Ours} & \multirow{2}{*}{V} &  & 5.8 &  & \textbf{2-bit} & 93.61 \\
 &  &  & 8.3 &  & \textbf{3-bit} & 96.87\\ \bottomrule\\
\end{tabular}
\caption{Comparison across other quantized/integerized models}
\label{tab:acc}
\vspace{-8mm}
\end{table}

Our low-bit integerized model enables integer-only inference while maintaining minimal accuracy loss compared to Q-ViT. This bridges the gap between low-bit quantized models and 8-bit integerized models, offering a more efficient inference for low-bit models.

\section{Conclusions}

In this paper, we introduce an integerization algorithm that reduces the computational cost of inference by operator reordering. 
Our approach enables a mix of low-bit matrix operations for efficiency and high-precision computations where necessary. To validate our method, we implement a systolic-based dataflow on FPGAs, demonstrating that low-bit operations result in lower per-PE power consumption as an efficient inference solution for low-bit quantized models.

% \section*{References}
\bibliographystyle{IEEEtran}
\bibliography{references}

\end{document}